\documentclass[runningheads]{llncs}

\usepackage{eccv}

\usepackage{eccvabbrv}

\usepackage{graphicx}
\usepackage{booktabs}
\usepackage{float}
\usepackage{bm}
\usepackage[accsupp]{axessibility}  

\usepackage{hyperref}

\usepackage{orcidlink}

\begin{document}

\title{DreamUV: Unwrap Artist-like UV by End-to-End Flow Matching} 


\author{Quanyuan Ruan\inst{1} \and
Jiabao Lei\inst{2} \and
Xingyi Du\inst{3} \and
Xifeng Gao\inst{3}
}

\authorrunning{Q. Ruan et al.}

\institute{South China University of Technology \and
School of Data Science, The Chinese University of Hong Kong, Shenzhen\and
Lightspeed
}


\maketitle

\begin{abstract}
UV parameterization is a fundamental step in 3D content creation, yet producing production-ready UV layouts remains challenging due to the gap between geometric distortion objectives and the stylistic preferences of professional artists.
While classical methods optimize handcrafted energy functions, artist-authored UVs exhibit structural patterns such as straightened seams, axis-aligned islands, and flexible interior deformation, properties that are difficult to explicitly formulate.

In this work, we present \textbf{DreamUV}, an end-to-end learning framework that formulates UV unwrapping as a generative Flow Matching problem.
Rather than predicting a single optimal parameterization, DreamUV learns a mesh-conditioned transport process that maps noise samples to a distribution of artist-like UV layouts.
To reflect real-world authoring practices, we introduce a \emph{boundary-aware training} strategy that prioritizes seam geometry, and a \emph{Model-in-the-Loop Finetuning} (MITL) scheme that explicitly accounts for discretization errors during sampling and stabilizes transport dynamics under heterogeneous supervision.

We evaluate DreamUV on a large-scale dataset of professionally authored UV layouts.
Experiments demonstrate that our method produces significantly straighter boundaries and tighter axis-aligned islands than both classical and learning-based baselines, while maintaining competitive distortion metrics.
Qualitative results and a user study with professional artists further confirm that DreamUV generates UV layouts that are not only valid, but aligned with practical production requirements.
\end{abstract}

\section{Introduction}
\label{sec:intro}

UV parameterization is a fundamental component of modern 3D content creation pipelines, providing a mapping from surface geometry to a planar domain, thereby enabling a wide range of applications including texture mapping, detail transfer, remeshing, morphing, and rendering\cite{floater2005surface}.
Despite decades of research, generating production-ready UV layouts remains challenging, as practical requirements in artist workflows often diverge from the mathematical objectives pursued by classical parameterization methods.

Traditional algorithms\cite{levy2002least,sheffer2005abf++} focus on preserving geometric properties such as conformality and equiareality, and thus formulate parameterization as an energy minimization problem with a handcrafted distortion metric, yielding only a single solution after optimization. 
While sensible from a purely geometric standpoint, these methods often produce irregular islands, curved seams, and suboptimal axis alignment efficiency. 
In contrast, artist-authored UV parameterizations exhibit strong structural regularities: seams are frequently straightened, boundaries are axis-aligned, and interior regions are permitted to stretch, ensuring that the overall layout remains visually clean with minimal distortion.
Artist-designed parameterizations are also usually diverse and non-deterministic, reflecting an individual's preferences. 
Such artistic preferences are therefore difficult to explicitly formulate and encode into the energy optimization objective. To address these challenges, our work adopts a data-driven approach instead of relying on handcrafted priors.

\begin{figure}[!t]
    \centering
    \includegraphics[width=1.0\linewidth]{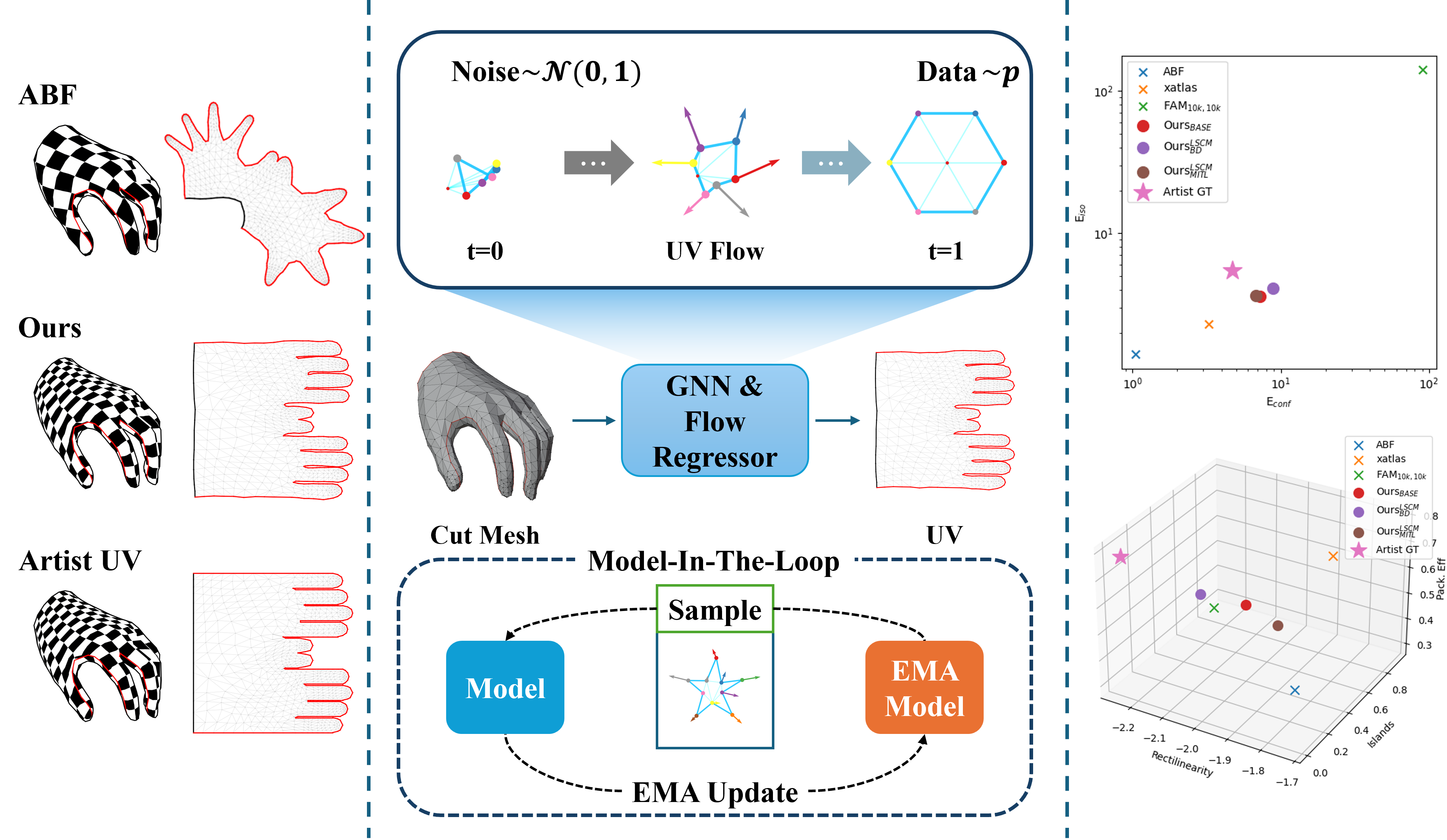}
    \caption{\textbf{DreamUV: Artist-Style UV Layout Generation via Flow Matching}
    \textbf{Left:} Qualitative comparison under identical seams. Our method generates cleaner boundaries and more structured layouts than the conventional Angle-Based Flattening (ABF), closely matching artist-created ground-truth UVs.
    \textbf{Middle:} Flow Matching formulation. Node coordinates are transported from Gaussian noise in UV space to the target parameterization via a learned UV flow field, implemented with a GNN-based flow regressor.
    \textbf{Bottom:} Model-in-the-loop training. Generated UV samples are reused during optimization, with an exponential moving average (EMA) model stabilizing training and mitigating error accumulation.
    \textbf{Right:} Multi-metric distribution comparison. Our method’s layout distribution aligns significantly closer to the artist ground-truth (star) target distribution than existing approaches, demonstrating superior structural matching to professional parameterizations.}
    \label{fig:teaser}
\end{figure}

Our study is grounded in a large-scale corpus of professionally authored UV layouts created by experienced technical artists. Unlike UVs produced by mathematical minimization or synthetic procedures, these layouts reflect real production choices, such as intentional boundary straightening and higher axis alignment efficiency, which prioritize usability over strict distortion optimality\cite{artuv}. This artist-authored data naturally encodes unwrapping preferences that reflect desired patterns and regularities otherwise difficult to explicitly formulate.

Since different artists have different unwrapping preferences, there naturally exist multiple sensible solutions for the same input. Inspired by this perspective, rather than modeling the problem deterministically, we formulate UV unwrapping as sampling from an implicit, mesh-conditioned distribution, thereby inherently introducing stochasticity to ensure result diversity.

Flow Matching is a simple yet effective generative modeling framework well-suited for generating diverse outputs\cite{lipman2022flow}.
Unlike classical optimization-based methods, Flow Matching learns a global velocity field whose integral curves lead to valid samples\cite{lipman2022flow, chen2018neural}, making it particularly effective for modeling distributions of plausible UV unwrappings rather than a single deterministic solution. 
In this paper, we build our approach upon an end-to-end Flow Matching framework operating on the mesh domain that learns to generate diverse artist-like UV parameterizations from random initialization.

However, directly applying Flow Matching to UV unwrapping poses several challenges. 
UV seams induce discontinuities, and small errors near boundaries can lead to severe visual artifacts. 
Moreover, while Flow Matching is theoretically grounded in continuous-time dynamics, practical inference relies on discrete numerical integration, leading to a potential mismatch between training objectives defined on instantaneous velocities and the accumulated behavior of sampled trajectories.
To address these challenges, we present \textbf{DreamUV}, a learning-based UV unwrapping framework with three major contributions:
\begin{itemize}
    \item \textbf{A Flow Matching formulation for UV unwrapping}, which reframes UV generation as a data-driven learning approach that employs a mesh-conditioned transport process from noise to artist-authored UV parameterizations, enabling the generation of artistic yet diverse solutions.
    \item \textbf{Boundary-aware training}, which explicitly prioritizes UV seam vertices during learning, reflecting the dominant role of boundary geometry in artist workflows and encouraging straight, well-aligned island boundaries.
    \item \textbf{Model-in-the-Loop Finetuning (MITL)}, a sampling-aware training strategy that exposes the model to its own discrete integration behavior, mitigating error accumulation and stabilizing transport dynamics under boundary-focused supervision.
\end{itemize}

Extensive experiments on a large-scale artist-authored dataset demonstrate that DreamUV generates results with significantly straighter seams and higher axis-aligned bounding box (AABB) occupancy ratios than classical and learning-based baselines, while maintaining competitive distortion metrics.
Qualitative results and a user study further confirm that our method produces UVs that are not only plausible but also aligned with practical production requirements.

\section{Related Work}

\paragraph{Classical UV parameterization.}
Classical UV unwrapping is rooted in geometric surface parameterization, which maps a 3D mesh to a 2D domain while preserving desirable geometric properties. Most traditional methods first cut the mesh into one or more topological disks, fix boundary loops onto planar curves, and then solve a sparse linear system to obtain UV coordinates \cite{floater2005surface}. Widely adopted solvers in artist-driven pipelines include Least-Squares Conformal Maps (LSCM) and Angle-Based Flattening (ABF/ABF++)~\cite{levy2002least,sheffer2005abf++,rabinovich2017scalable,jiang2017simplicial}. Fixing boundary constraints guarantees injectivity but often introduces significant distortion, especially near seams; free-boundary formulations such as ABF++ partially alleviate this by relaxing boundary conditions to better preserve angles~\cite{sheffer2005abf++}. Despite their effectiveness, these methods rely on high-quality manifold meshes and typically require careful seam placement, preprocessing, or remeshing when applied to complex, noisy, or reconstruction-derived geometry\cite{sorkine2002bounded,li2018optcuts}.

\paragraph{Learning-based UV unwrapping.}
Recent years have seen growing interest in learning-based approaches to surface parameterization. Early neural models such as FoldingNet \cite{yang2018foldingnet} and AtlasNet \cite{groueix2018papier} learn to deform a canonical 2D domain into 3D shapes, but they are primarily designed for shape reconstruction and do not explicitly enforce valid UV mappings or distortion control. More recent works aim to directly learn surface-to-UV mappings. Nuvo \cite{srinivasan2024nuvo} represents a continuous UV function using an MLP defined over 3D points and trains only on visible surface samples, making it robust to imperfect or non-manifold reconstructions such as NeRF outputs. 
The Flatten Anything Model (FAM) \cite{zhang2024flatten} proposes an unsupervised framework that mimics the classical pipeline with learnable components for cutting, flattening, and wrapping, enabling free-boundary parameterization on point clouds without explicit pre-cutting.
FlexPara \cite{zhao2025flexpara} further introduces a flexible neural parameterization framework that improves adaptability across diverse surface geometries.
PartUV\cite{wang2025partuv} further incorporates learned part decomposition, segmenting shapes into semantic components and parameterizing each part separately to produce fewer, better-aligned UV charts with reduced distortion. ArtUV \cite{artuv} learns to refine an initial UV parameterization by predicting per-vertex UV offsets, transforming a coarse geometric map into an artist-style layout. By directly modeling the stylistic characteristics of professional UV designs, it produces cleaner boundaries, improved chart alignment, and more structured layouts compared to purely geometric methods.

Other learning-based schemes focus on local parameterizations or alternative surface representations. Methods such as DiffSR aggregate locally learned surface parameterizations to reconstruct global geometry, imposing stronger consistency constraints than simple neural deformation models~\cite{zhang2024parapoint}. RegGeoNet and Flattening-Net convert irregular point clouds into regular geometry images (PGIs), approximating local 3D-to-2D flattening while preserving neighborhood structure~\cite{zhang2022reggeonet, zhang2023flattening}. However, these approaches do not explicitly optimize for global UV validity, seam quality, or distortion measures central to UV unwrapping. In contrast, our method directly formulates UV unwrapping as a generative flow problem, learning a global mapping from noise to UV coordinates with explicit boundary weighting and sampling-aware optimization.

\paragraph{Flow-based and correspondence-driven mappings.}
Flow-based models and methods based on correspondence provide a complementary perspective by learning continuous, often invertible mappings between geometric domains. FUSE~\cite{olearo2025fuse} employs continuous normalizing flows to learn bijective correspondences between 3D shapes, representing shapes as distributions and transporting points through learned flows. While primarily designed for shape correspondence, FUSE reports promising preliminary results on UV mapping tasks. Our approach is conceptually related but differs fundamentally in scope and formulation: instead of learning correspondences between shapes, we apply Flow Matching directly to the UV unwrapping problem. We learn per-vertex velocity fields that transport samples from a simple prior to valid UV embeddings, with dedicated mechanisms to handle seams and to stabilize discrete sampling dynamics (Sec.~\ref{sec:method}).

\section{Method}
\label{sec:method}

In this section, we present an end-to-end, Flow Matching-based approach for UV unwrapping. We begin with a brief formal formulation that situates Flow Matching within the framework of continuous normalizing flows (CNFs) and defines the equations used in subsequent subsections. We then describe three technical components: (i) the basic Flow Matching-based formulation for UV unwrapping; (ii) a boundary-focused weighting mechanism that concentrates learning on seam geometry; and (iii) a model-in-the-loop fine-tuning scheme that closes the gap between training and sampling dynamics.

\subsection{Flow Matching: background and notation}

Continuous normalizing flows (CNFs)~\cite{chen2018neural} model a density path ${p_t},\ {t\in[0,1]}$ by an instantaneous velocity field $v_t(x)$ that transports samples according to the ordinary differential equation
\begin{equation}\label{eq:ode}
\frac{\mathrm{d}x}{\mathrm{d}t} = v_t(x), \qquad x(0)=x_0 \sim p_0,
\end{equation}
so that $x(1)\sim p_1$ under perfect integration. Flow Matching (FM)~\cite{lipman2022flow} avoids explicit likelihood computation by training a parametric velocity field $v_\theta(x,t)$ to match a chosen conditional vector field $u_t(x\mid x_1)$ that defines trajectories between source and target samples. Concretely, given a simple prior $p_0$ (\textit{e.g.}, isotropic Gaussian) and the data distribution $p_1$ of ground-truth UV coordinates, FM minimizes a regression loss between $v_\theta$ and $u_t$ under the joint $(x_0,x_1)\sim p_0\times p_1$ and uniformly sampled time $t\sim\mathcal{U}[0,1]$.

\begin{figure}
    \centering
    \includegraphics[width=0.9\linewidth]{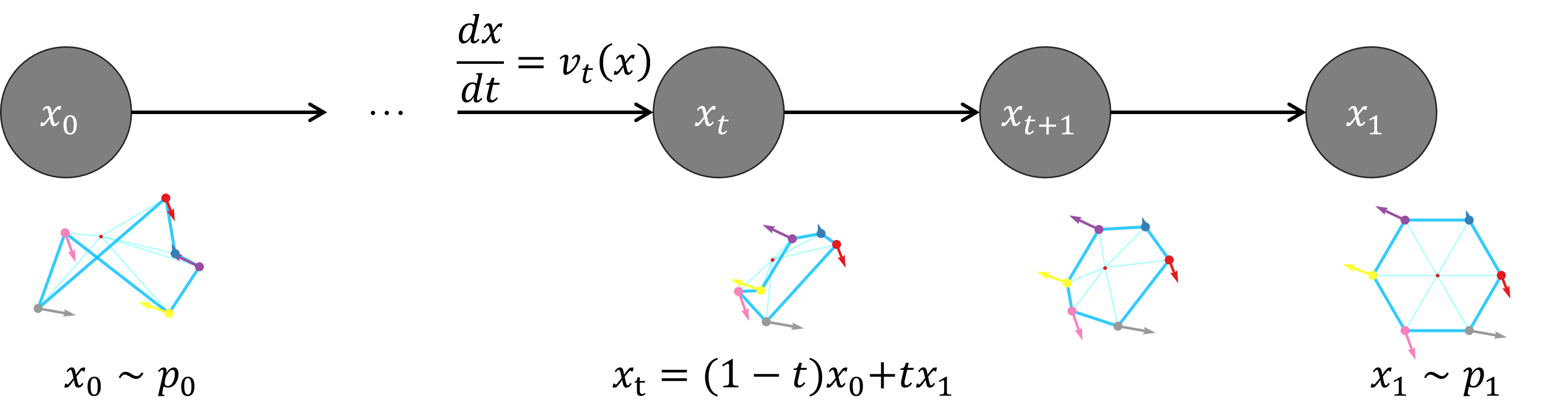}
    \caption{\textbf{Illustration of the Flow Matching process for UV unwrapping.} The model integrates an ODE~\cite{chen2018neural} from $t=0$ to $t=1$, transporting noisy initial coordinates $x_0 \sim p_0$ (left) along linear trajectories to valid ground-truth UV coordinates $x_1 \sim p_1$ (right). The arrows on the vertices represent the constant velocity field $u_t = x_1 - x_0$ pointing toward the target.}
    \label{fig:placeholder}
\end{figure}

A natural and widely-used choice for $u_t$ is the Optimal Transport (OT)~\cite{lipman2022flow} conditional which defines straight-line (linear) trajectories from $x_0$ to $x_1$:
\begin{align}
x_t &= (1-t)x_0 + t x_1\label{eq:linear_interp}\\
u_t(x_t\mid x_1) &= x_1 - x_0.\label{eq:ot_vector}
\end{align}
Under this choice the target velocity is constant along the trajectory and points from the source to the target. Training $v_\theta$ to match $u_t$ yields a vector field whose integral curves move noise samples toward valid UV coordinates. The canonical Flow Matching objective is therefore the expected mean squared error (MSE)
\begin{equation}\label{eq:fm_basic}
\mathcal{L}_{FM}^{\text{vanilla}}(\theta)
= \mathbb{E}_{t\sim\mathcal{U}[0,1],x_0\sim p_0,x_1\sim p_1}
\big[|v_\theta(x_t,t) - (x_1-x_0)|^2\big].
\end{equation}

Below we describe adaptations of this objective that are tailored to the UV unwrapping problem: (i) spatial weighting to emphasize boundary correctness, and (ii) a model-in-the-loop training phase that accounts for errors accumulated during discrete sampling.

\subsection{Flow Matching UV Unwrap}
We formulate UV unwrapping as learning a mapping that transports prior samples $x_0\sim p_0$ to UV coordinates $x_1\sim p_1$ using the FM framework above. During training we sample $t$, construct $x_t$ by linear interpolation (Eq.~\ref{eq:linear_interp}), and regress $v_\theta(x_t,t)$ to the OT target $x_1-x_0$ (Eq.~\ref{eq:ot_vector}). 

\subsection{Boundary-focus}

UV quality is particularly sensitive to seam placement: small errors near UV island boundaries often lead to visually salient distortions in texture alignment and packing.
In contrast, inaccuracies in interior regions are typically far less noticeable and can often be corrected implicitly through smooth interpolation.
However, the standard Flow Matching MSE objective (Eq.~\ref{eq:fm_basic}) treats all vertices equally, which tends to under-emphasize seam fidelity.

This mismatch becomes especially apparent when considering how UV layouts are authored in practice.
Professional artists rarely manipulate all UV vertices uniformly; instead, they focus almost exclusively on adjusting and straightening UV boundaries, while allowing interior vertices to be filled in automatically using optimization-based methods such as LSCM\cite{levy2002least} or simple interpolation.
As a result, boundary geometry encodes the dominant structural intent of the UV layout, whereas interior regions primarily serve to smoothly propagate these constraints.

From a geometric perspective, boundary vertices correspond to regions where the mapping is weakly constrained and highly sensitive to perturbations.
Small deviations near seams may have limited influence on global distortion energies, yet they produce disproportionately large visual artifacts.
Interior vertices, on the other hand, are typically well-conditioned and can be reliably recovered once boundary conditions are fixed.

From a probabilistic viewpoint, standard Flow Matching assumes a uniform importance measure over the state space.
We relax this assumption by introducing a spatially varying weighting function that reflects the asymmetric role of boundary and interior vertices in artist-authored UV design.
Specifically, we redefine the training objective under a non-uniform measure that concentrates probability mass near boundary-induced singular regions.
This formulation encourages the learned vector field to explicitly model seam geometry, while allowing interior regions to be resolved implicitly through smooth transport dynamics, analogous to classical free-boundary parameterization methods.

Let $\mathcal{V}$ denote the set of mesh vertices and $\mathcal{V}_{\mathrm{boundary}}\subset\mathcal{V}$ the subset that lie on UV island boundaries.
We assign a scalar weight $w_i$ to each vertex $i\in\mathcal{V}$:
\begin{equation}
w_i =
\begin{cases}
w_{\mathrm{b}} & i \in \mathcal{V}_{\mathrm{boundary}},\\
w_{\mathrm{*}} & i \notin \mathcal{V}_{\mathrm{boundary}}.
\end{cases}
\end{equation}
with $w_{\mathrm{b}}\gg w_{\mathrm{*}}$ (in practice $w_{\mathrm{b}}=1.0$, $w_{\mathrm{*}}=0.1$ unless stated otherwise).
Applying these weights to the Flow Matching objective yields the boundary-weighted velocity loss:
\begin{equation}\label{eq:fm_weighted}
\mathcal{L}_{FM}(\theta)
= \mathbb{E}_{t,x_0,x_1}\left[
\sum_{i\in\mathcal{V}}
w_i \times \big\lVert v_\theta(x_t,t)_i - (x_1-x_0)_i \big\rVert^2
\right],
\end{equation}
where the subscript $i$ indicates the vertex-specific components of the vector field.
This weighting scheme explicitly prioritizes seam correctness, while permitting less constrained, LSCM-like relaxation of interior regions.

\subsection{Model-in-The-Loop Finetuning}

\begin{figure}
    \centering
    \includegraphics[width=0.9\linewidth]{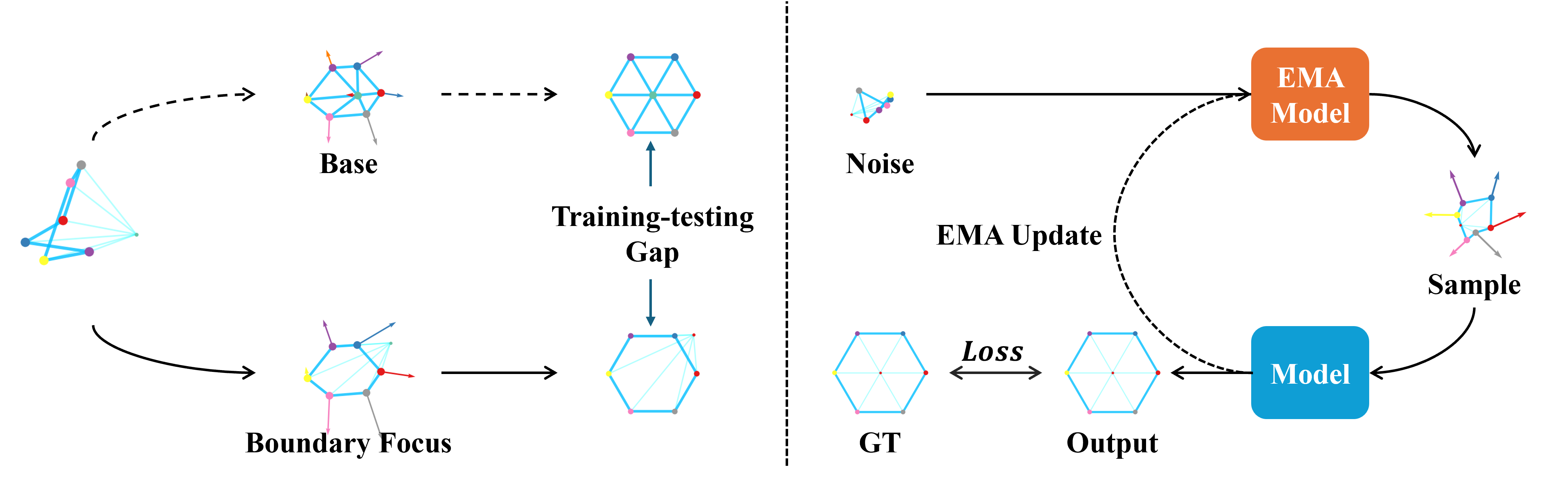}
    \caption{\textbf{Illustration of the Model-in-the-Loop Finetuning (MITL) Pipeline.} 
    \textbf{Left}: Comparison between the Base model (balanced supervision on boundary and interior points) and Boundary Focus model (weaker supervision on interior points). Despite identical per-point error expectation in training, the Boundary Focus model suffers from amplified multi-step error accumulation during inference, leading to a severe training-testing gap. 
    \textbf{Right}: The MITL framework to mitigate this gap: An EMA model simulates inference-time cumulative errors (Sample) via forward prediction; the active model predicts final coordinates (Output) from perturbed states; a multi-step discrepancy loss against ground truth (GT) trains the model to correct accumulated errors, aligning training behavior with inference scenarios.}
    \label{fig:placeholder}
\end{figure}

Flow Matching provides theoretical guarantees in the continuous-time limit, where samples evolve under an ordinary differential equation governed by a learned velocity field.
At inference time, however, this continuous dynamics must be approximated using a finite-step numerical solver, inevitably introducing discretization errors that are not explicitly penalized during standard training.

This discrepancy is further amplified by our boundary-focused training scheme.
By construction, boundary vertices receive significantly stronger supervision, resulting in a higher signal-to-noise ratio (SNR) in their learned velocity estimates.
Interior vertices, in contrast, are weakly supervised and are expected to emerge implicitly through smooth propagation, leading to noisier velocity predictions.
While such asymmetric supervision faithfully reflects the distinct roles of boundary and interior regions in artist-authored UV layouts, it poses a critical challenge under discrete-time integration: noise in interior velocities can accumulate across steps and destabilize the transport dynamics.

Consequently, a distribution shift arises between the training objective, which supervises instantaneous velocities, and the sampling process, which relies on repeated discrete integration.
Inspired by~\cite{sam3dteam2025sam3d3dfyimages},
we propose a \emph{Model-in-the-Loop Finetuning} (MITL) strategy that explicitly enforces consistency between the learned velocity field and its induced discrete-time flow.
Rather than treating discretization errors as incidental, MITL exposes the model to its own integration behavior during training, enabling it to actively compensate for error accumulation caused by heterogeneous supervision.

Specifically, standard Flow Matching optimizes instantaneous velocity matching but does not penalize deviations that compound over multiple integration steps.
MITL bridges this train-test gap by simulating sampling dynamics during training and forcing the model to correct multi-step errors, thereby restoring stability under boundary-focused supervision.

Given a sampled pair $(\bm{x}_0, \bm{x}_1)$ and times $0 \le t_1 < t_2 < 1$, the finetuning step proceeds as follows:
\begin{enumerate}
\item \textbf{Look-ahead (predict with EMA):}
We form the intermediate state $x_{t_1} = (1 - t_1)x_0 + t_1 x_1$ and use an exponential moving average model $v_{\theta_{\mathrm{EMA}}}$ to predict a forward step:
\begin{equation}\label{eq:lookahead}
\hat{x}_{t_2} = x_{t_1} + v_{\theta_{\mathrm{EMA}}}(x_{t_1}, t_1)\cdot (t_2 - t_1).
\end{equation}
The EMA model approximates the model behavior at inference time and thus simulates realistic deviations induced by discretized integration.

\item \textbf{Correction (active model):}
Starting from the perturbed state $\hat{x}_{t_2}$, we use the current active model $v_{\theta}$ to predict the remaining trajectory to the final time $t=1$:
\begin{equation}\label{eq:correction}
\hat{x}_1 = \hat{x}_{t_2} + v_{\theta}(\hat{x}_{t_2}, t_2)\cdot (1 - t_2).
\end{equation}
This produces a multi-step, model-driven estimate $\hat{x}_1$ of the final UV coordinates.
\end{enumerate}

Training then minimizes the discrepancy between the multi-step prediction $\hat{x}_1$ and the ground-truth target $x_1$. 
This multi-step objective, enforcing vertex-wise fidelity after the look-ahead and correction steps, uses the boundary weights $w_i$ from Eq.~\ref{eq:fm_weighted}: 
\begin{equation}
\label{eq:mitl_loss} 
\mathcal{L}_{\mathrm{MITL}}(\theta) = \mathbb{E}_{t_1,t_2,x_0,x_1}\left[ \sum_{i\in\mathcal{V}} w_i \times \big\lVert \hat{x}_{1,i} - (x_1)_i \big\rVert^2 \right], 
\end{equation} 
where $\hat{x}_1$ is obtained via Eqs.~\ref{eq:lookahead}--\ref{eq:correction}. 

In practice, we first train the model using the boundary-weighted Flow Matching objective (Eq.~\ref{eq:fm_weighted}) to learn stable instantaneous velocities, followed by the proposed model-in-the-loop finetuning that optimizes Eq.~\ref{eq:mitl_loss}. 
This two-stage procedure yields velocity fields that not only match OT trajectories locally, but also produce stable and geometry-preserving UV samples under the discrete sampling dynamics used at inference.

\section{Experiments}
\label{sec:experiments}

We evaluate \textbf{DreamUV} with a focus on its ability to generate ``Artist-like'' parameterizations, specifically, UV layouts that balance geometric low-distortion with the structural regularity preferred in professional modeling pipelines.

\subsection{Experimental Setup}

\paragraph{Datasets.}
We introduce the \textbf{Lightspeed Games UV dataset}, consisting of 359,301 meshes with valid UV connectivity.
All ground-truth UV parameterizations are manually authored by professional 3D artists.
Compared to UVs produced by purely geometric minimization, these artist-created layouts exhibit consistent structural characteristics, including intentionally straightened seams and axis-aligned island boundaries.


Unlike synthetic or energy-minimized UVs, the layouts in our dataset reflect real production decisions made in professional content creation pipelines.
As such, they provide a strong supervision signal for learning stylistic and structural priors that are difficult to encode explicitly through analytic parameterization energies.

\paragraph{Models.}

Given vertex positions, surface normals, noisy UV coordinates, and boundary indicators, the model first encodes geometric information using spherical harmonics on normals and concatenates it with positional, UV, and boundary embeddings. A sinusoidal time embedding is injected throughout the network via adaptive normalization, enabling conditional flow prediction at different time steps. The backbone consists of multiple residual graph blocks with pre-normalization, combining EdgeConv-based\cite{wang2019dynamic} local geometric aggregation and multi-head attention-based graph convolutions to capture both local structural cues and long-range dependencies. To incorporate global context, the network performs attention-based graph pooling to obtain a graph-level representation, which is fused back into node features before the prediction head. Finally, a linear projection outputs 2D UV flow vectors per vertex, followed by a per-graph zero-mean normalization to satisfy flow consistency. This architecture integrates geometric priors, temporal conditioning, and multi-scale graph reasoning in a unified framework for UV flow regression.

\paragraph{Metrics.}
Beyond standard distortion measures~\cite{shay2022dataset}, we use the following metrics to quantify \emph{artist-like} UV characteristics:
\begin{itemize}
    \item \textbf{Conformal Distortion ($E_{\mathrm{conf}}$)} measures angular distortion. For each triangle, let $\sigma_1 \ge \sigma_2$ be the singular values of the Jacobian mapping UV to 3D. We define
    \begin{equation}
        E_{\mathrm{conf}} = \frac{\sigma_1}{\sigma_2}.
    \end{equation}
    $E_{\mathrm{conf}}=1$ indicates perfect angle preservation; larger values imply stronger anisotropy.

    \item \textbf{Isometric Distortion ($E_{\mathrm{iso}}$)} measures local metric stretch/compression:
    \begin{equation}
        E_{\mathrm{iso}} = \max\left(\sigma_1, \frac{1}{\sigma_2}\right).
    \end{equation}
    $E_{\mathrm{iso}}=1$ is perfectly isometric; values $>1$ indicate expansion or contraction. For both distortion metrics, we report the area-weighted mean over all valid faces.

    \item \textbf{Self-Overlap ($E_{\mathrm{overlap}}$)} quantifies global non-injectivity (fold-overs) as the fraction of redundant UV area:
    \begin{equation}
        E_{\mathrm{overlap}} = \frac{\sum_i A(T_i) - A\!\left(\bigcup_i T_i\right)}{\sum_i A(T_i)},
    \end{equation}
    where $A(T_i)$ is the signed UV area of triangle $i$, and $A(\bigcup_i T_i)$ is the area of the union of all UV triangles. $E_{\mathrm{overlap}}=0$ indicates no overlap.

    \item \textbf{Boundary Rectilinearity ($S_{\mathrm{rect}}$)} measures how axis-aligned and piecewise-linear UV boundaries are. We split boundary loops into chains at high-curvature vertices (turning angle $>45^\circ$). For each chain, we fit a line using PCA and compute the vertex-to-line RMSE. $S_{\mathrm{rect}}$ is the length-weighted average RMSE; lower is better.

    \item \textbf{UV Islands ($N_{\mathrm{islands}}$)} counts connected components in UV space. We build a graph with UV vertices as nodes and face edges as connections; each connected component defines one island. Fewer islands typically imply fewer seams and a more artist-friendly layout.

    \item \textbf{Packing Efficiency ($\eta_{\mathrm{pack}}$)} measures texture-space utilization, defined as the ratio between the total island area and the area of the tightest axis-aligned bounding box enclosing the UV layout.
\end{itemize}

\begin{figure}[!t]
    \centering
    \includegraphics[width=0.9\linewidth]{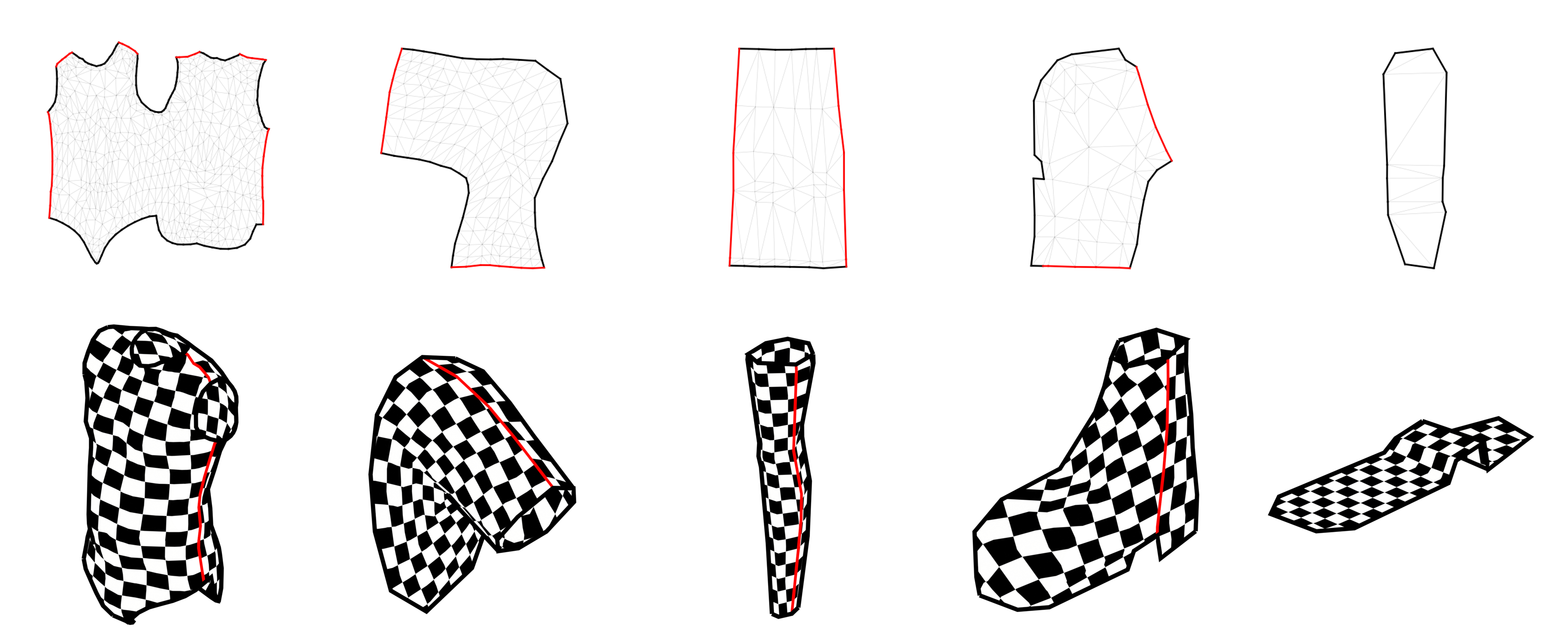}
    \caption{\textbf{Zero-shot inference.} We use meshes from SMPL and Sketchfab to generate the output data, demonstrating our zero-shot capability.}
    \label{fig:placeholder}
\end{figure}

\subsection{Model Configuration}

We adopt the same training and evaluation protocol as in the rest of the paper and report all results on the Artist-Authored validation split. We denote model size as \texttt{<depth>-layers / <width>-hidden}; for example, \texttt{24$\times$1024} corresponds to 24 residual graph blocks with hidden width 1024.

\paragraph{Ablation design.}
We ablate model scaling along two axes: (i) increasing depth (8, 12, 24 layers) while keeping width fixed at 1024, and (ii) increasing width (256, 512, 1024) while keeping depth fixed at 24. For each configuration, we report the approximate parameter count and the full set of quality metrics, including conformal and isometric distortion, overlap rate, boundary rectilinearity, island count, and packing efficiency. All numbers are area-weighted averages on the Artist-Authored validation set.

\paragraph{Discussion.}
Table~\ref{tab:ablation_scaling} suggests that widening the model is more impactful than simply stacking more layers. Increasing depth at fixed width does not yield consistent improvements: while deeper models can reduce overlap and slightly improve boundary rectilinearity, distortion metrics do not monotonically improve and may degrade at 24 layers. In contrast, at a fixed depth of 24, increasing width leads to more reliable gains across key metrics (notably rectilinearity and packing efficiency) and is therefore the more effective scaling direction in our setting.

\begin{table}[!t]
\small
\centering
\caption{\textbf{Ablation over depth and width.} Values are measured on the Lightspeed Games UV Test Set. 
}
\label{tab:ablation_scaling}
{
\begin{tabular}{lrrrrrrr}
\toprule
\textbf{Model} & \textbf{Params (M)} & \textbf{$E_{\mathrm{conf}}$} $\downarrow$ & \textbf{$E_{\mathrm{iso}}$} $\downarrow$ & \textbf{Overlap (\%)} $\downarrow$ & \textbf{$S_{\mathrm{rect}}$} $\downarrow$ & \textbf{Islands} $\downarrow$ & \textbf{$\eta_{\mathrm{pack}}$} $\uparrow$ \\
\midrule
8 × 1024  &  41.6 M & 6.3600 & 3.3708 & 0.0004 & 0.1813 & 1.0000 & 0.7046  \\
12 × 1024 &  60.8 M & 6.1793 & 3.3511 & 0.0002 & 0.1656 & 1.0000 & 0.6857 \\
24 × 1024 & 118.0 M & 7.2607 & 3.6059 & 0.0000 & 0.1549 & 1.0000 & 0.7449 \\
\midrule
24 × 256  &   7.7 M & 6.5991 & 3.4381 & 0.0007 & 0.2158 & 1.0000 & 0.7245 \\
24 × 512  &  30.0 M & 6.5761 & 3.3735 & 0.0007 & 0.1746 & 1.0000 & 0.7200 \\
24 × 1024 & 118.0 M & 7.2607 & 3.6059 & 0.0000 & 0.1549 & 1.0000 & 0.7449  \\
\bottomrule
\end{tabular}
}
\end{table}

\subsection{Comparison with SOTA and Classical Methods}

We compare DreamUV with classical optimization-based UV unwrapping methods (ABF~\cite{sheffer2005abf++}, ABF++~\cite{sheffer2005abf++}, and xatlas) as well as a recent learning-based baseline, FAM~\cite{zhang2024flatten}.

\begin{table}[!b]
\centering 
\caption{\textbf{Quantitative comparison on the Lightspeed Games UV Test Set.} For FAM, we denote the configuration as FAM$_{\text{number of sampled points}}^{\text{optimization steps}}$, where the subscript indicates the number of sampled points and the superscript denotes the number of optimization steps.
} 
\label{tab:main_results} 
{%
\begin{tabular}{lcccccc} 
\toprule 
\textbf{Method} & \textbf{$E_{\mathrm{conf}}$} $\downarrow$ & \textbf{$E_{\mathrm{iso}}$} $\downarrow$ & \textbf{Overlaps (\%)} $\downarrow$ & $S_{\mathrm{rect}}$) $\downarrow$ & \textbf{Islands $\downarrow$} & $\eta_{\mathrm{pack}}$) $\uparrow$ \\ 
\midrule 
ABF    & 1.0522          & \textbf{1.4128} & 0.0002          & 0.1785          & \textbf{1.0000} & 0.4878 \\ 
ABF++  & \textbf{1.0517} & 1.4130          & 0.0002          & 0.1785          & \textbf{1.0000} & 0.4880 \\ 
xatlas & 3.2621          & 2.3090          & \textbf{0.0000} & \textbf{0.1495} & 2.5325          & 0.5937 \\ 
\midrule 
FAM$_{1k}^{1k}$   & 172.5905 & 293.9393 & 0.0022 & \textbf{0.0588} & 4.0791 & 0.2159 \\ 
FAM$_{10k}^{1k}$  & 125.5837 & 188.2207 & 0.0028 & 0.1046          & 2.9681 & 0.2615 \\ 
FAM$_{10k}^{10k}$ &  89.9502 & 140.3959 & 0.0010 & 0.1048          & 2.3259 & 0.2917 \\ 
\textbf{Ours}$_\textrm{BASE}$ & 7.2607 & 3.6059 & \textbf{0.0000} & 0.1549 & \textbf{1.0000} & \textbf{0.7449} \\ 
\textbf{Ours}$_\textrm{BD}$ & 10.8779 & 5.4106 & 0.0241 & 0.1302 & \textbf{1.0000} & 0.7386 \\ 
\textbf{Ours}$_{\textrm{MITL}}$ & \textbf{6.6260}& \textbf{3.4879}& 0.0003 & 0.1666 & \textbf{1.0000} & 0.7077 \\ 

\midrule 
\textit{Artist GT} & 4.6906 & 5.5002 & 0.0000 & 0.1064 & 1.0000 & 0.7872 \\ 
\bottomrule 
\end{tabular}%
}
\end{table}

\begin{figure}[!t]
    \centering
    \includegraphics[width=0.9\linewidth]{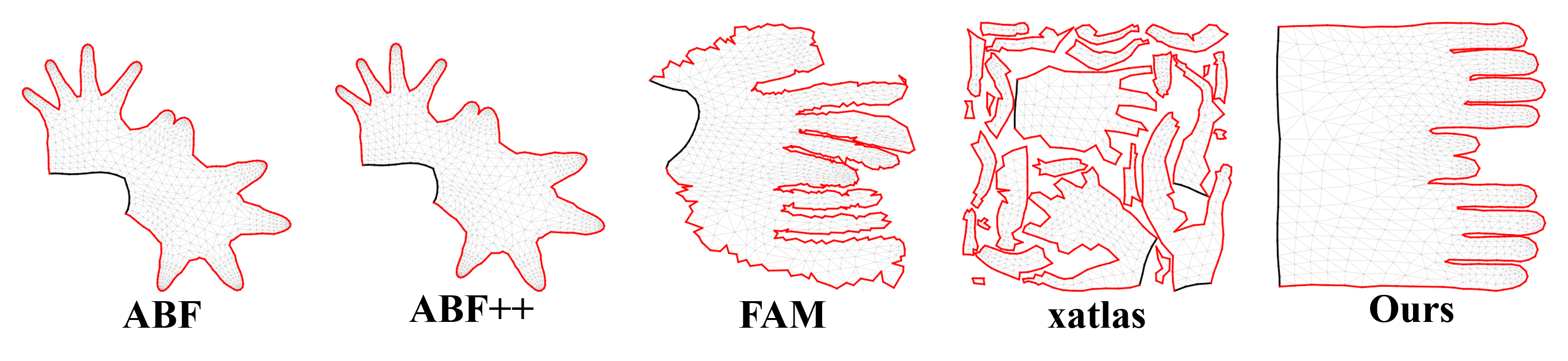}
    \caption{\textbf{Straight-seam comparison of UV layout.} ABF/ABF++ yield smooth single-chart layouts but with boundary distortion; FAM produces irregular, tangled triangulation due to unconstrained seams; xatlas forms clean seams but highly fragmented charts. Our method generates a compact, well-structured atlas with coherent, straighter boundaries and reduced fragmentation.}
    \label{fig:straight_seams}
\end{figure}

\begin{figure}[!b]
    \centering
    \includegraphics[width=0.9\linewidth]{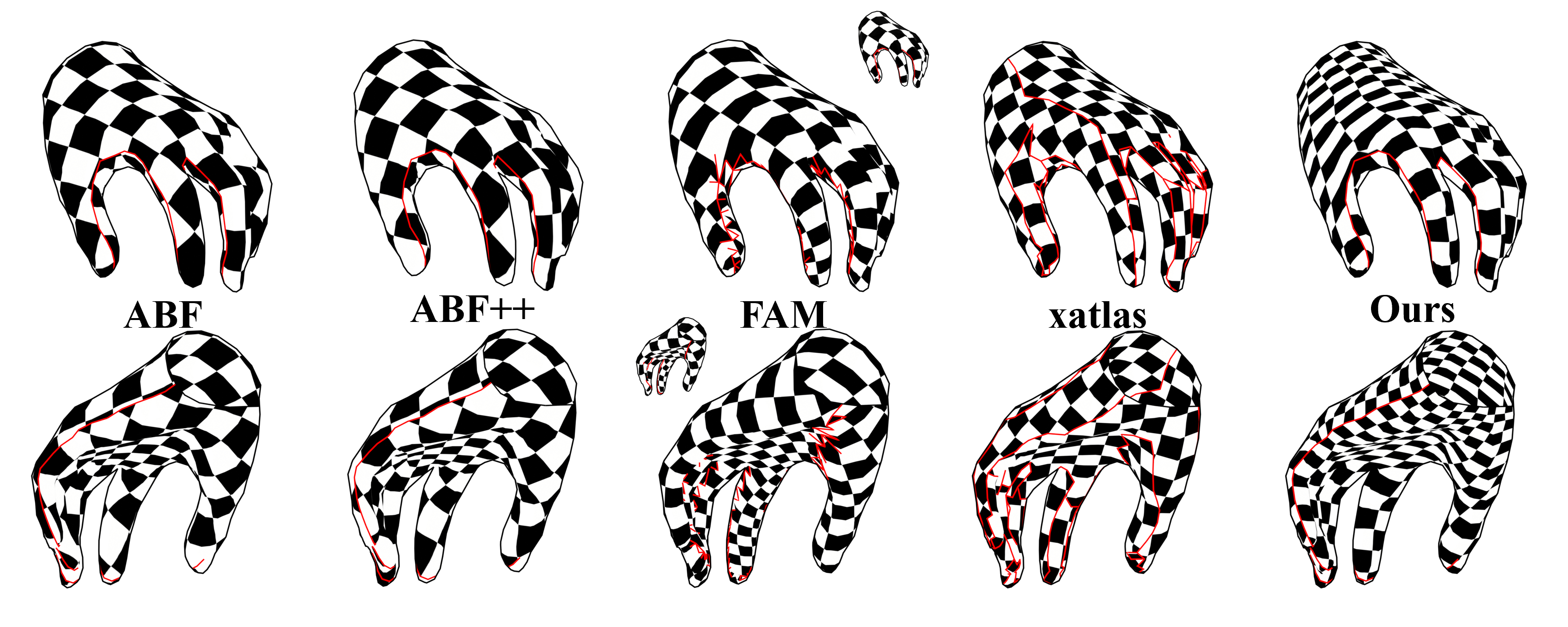}
    \caption{\textbf{Checkerboard texture visualization on a 3D hand.} Red curves mark seam locations. Our parameterization better preserves grid alignment along seams, while ABF++ introduces rotation/shearing and FAM induces irregular discontinuities that are not constrained to mesh edges.} 
    \label{fig:texture_grid}
\end{figure}

Table~\ref{tab:main_results} shows that optimization baselines (ABF/ABF++ and xatlas) achieve very low distortion, but produce less axis-rectilinear boundaries and markedly worse packing than artist-authored UVs. ABF/ABF++ minimize $E_{\mathrm{conf}}$/$E_{\mathrm{iso}}$ yet have higher boundary non-rectilinearity ($S_{\mathrm{rect}}$) and low $\eta_{\mathrm{pack}}$; xatlas improves $\eta_{\mathrm{pack}}$ and removes overlaps, but increases distortion and chart fragmentation (more islands). 
FAM learns a UV parameterization from a continuous point set via self-supervision, avoiding explicit mesh-edge discretization, but its predicted boundaries/seams are not guaranteed to lie on mesh edges. We therefore infer a discrete mesh seam by interpolating from FAM’s UVs. When the inferred seam crosses adjacent edges, it can create jagged, staircase artifacts, often requiring remeshing to better match the learned seam (see the FAM inset in Fig.~\ref{fig:texture_grid}).
DreamUV instead targets artist-like layout statistics: our variants keep (near-)zero overlaps and only one island, substantially boost packing efficiency toward Artist GT, and yield more rectilinear boundaries than classical optimization methods.

\subsection{Qualitative Analysis: The "Artist-like" Look}

\paragraph{Boundary Straightening.}

A key requirement for production UVs is that seams (e.g., along a pant leg or a cylinder cap) should map to straight lines to facilitate texture painting and reduce aliasing. Figure~\ref{fig:straight_seams} demonstrates this capability. 

\paragraph{Texture Alignment.}

In Figure~\ref{fig:texture_grid}, we apply a checkerboard texture. On the DreamUV parameterization, the checkerboard lines remain parallel to the seam boundaries, a desirable property for hard-surface modeling and clothing, whereas ABF++ introduces rotation and shearing relative to the seams.


\subsection{User Study}

We conducted a blind A/B study with 12 professional 3D artists to measure preference among UV unwrapping methods. In each trial, participants saw two raw UV layouts side-by-side (DreamUV vs.\ a baseline) and answered:

\textit{“Which UV layout would you prefer to use for manual texturing?”}

\begin{figure}[!t]
    \centering
    \includegraphics[width=1.0\linewidth]{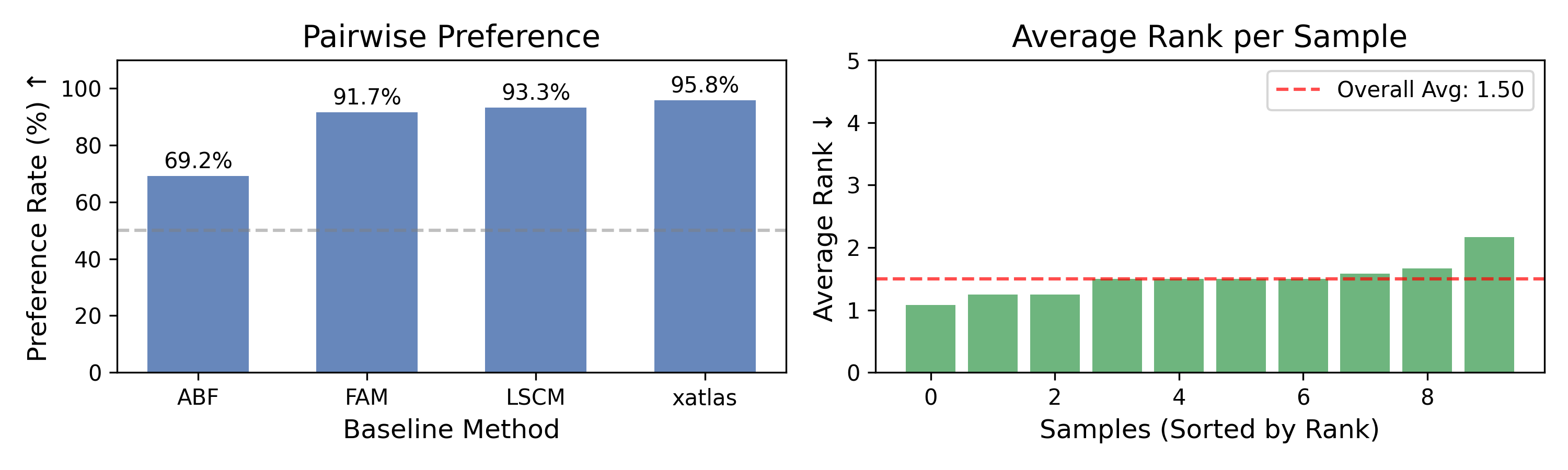}
    \caption{\textbf{User preference study results.} We asked users to choose between DreamUV and four classical UV parameterization methods in pairwise comparisons (left). DreamUV is preferred in most cases (69.2\%--95.8\% depending on the baseline). When ranking methods across samples (right; lower is better), DreamUV achieves an average rank of 1.50, indicating it is the overall top choice.}
    \label{fig:user_study}
\end{figure}

Baselines were ABF, FAM, LSCM, and xatlas. Order was randomized and method names were hidden. We report DreamUV’s win rate against each baseline (Fig.~\ref{fig:user_study}); DreamUV wins consistently across all comparisons. Artists cite better structural regularity and straighter boundaries, which ease manual texturing. 

\section{Conclusion}
DreamUV demonstrates that generative Flow Matching can go beyond mathematical energy minimization to capture the tacit stylistic rules of human artists. By learning from data, we produce UVs that are not just valid, but production-ready.

Despite these promising results, our approach has certain limitations. First, DreamUV operates directly on pre-cut meshes, relying on existing seams rather than predicting or generating the seam placements\cite{poranne2017autocuts,Yu_2018_ECCV,wang2020pie,bazazian2021edc,himeur2021pcednet,li2025auto}. Second, our method focuses exclusively on the unwrapping phase and does not address UV packing\cite{levy2002least,limper2018box,liu2019atlas,yang2023learning,vining2025fastatlas,wang2025partuv,artuv}. 
Integrating automatic seam generation and UV packing into an end-to-end pipeline presents an exciting avenue for future research.

%
%
\vspace{1cm}
\bibliographystyle{splncs04}
\bibliography{main}
\end{document}